\documentclass{article}
\usepackage{ijcai22}

\usepackage{times}

\usepackage{soul}
\usepackage{url}
\usepackage[hidelinks]{hyperref}
\usepackage[utf8]{inputenc}
\usepackage[small]{caption}
\usepackage{graphicx}
\usepackage{amsfonts}
\usepackage{amssymb,amsmath,amsthm}
\usepackage{booktabs}
\usepackage{algorithm}
\usepackage{algpseudocode}  
\usepackage{xcolor}
\usepackage{pifont}
\newcommand{\cmark}{\ding{51}}%
\newcommand{\xmark}{\ding{55}}%
\usepackage{url} 
\DeclareFontFamily{OT1}{pzc}{}
\DeclareFontShape{OT1}{pzc}{m}{it}{<-> s * [1.10] pzcmi7t}{}
\DeclareMathAlphabet{\mathpzc}{OT1}{pzc}{m}{it}
\newtheorem{theorem}{Theorem}
\newtheorem{corollary}{Corollary}

\title{Distribution-aware Margin Calibration for Semantic Segmentation in Images}

\author{
Litao Yu$^1*$\and
Zhibin Li$^2*$\and
Min Xu$^1$\and
Yongsheng Gao$^3$\and
Jiebo Luo$^4$ \And
Jian Zhang$^1$ \\
\affiliations
$^1$University of Technology Sydney\\
$^2$CSIRO\\
$^3$Griffith University\\
$^4$Rochester University\\
* Equally contributed. \\
\emails
\{Litao.Yu, Min.Xu, Jian.Zhang\}@uts.edu.au,
Zhibin.Li@csiro.au,
yongsheng.gao@griffith.edu.au,
jluo@cs.rochester.edu}
\begin{document}

\maketitle

\begin{abstract}
The Jaccard index, also known as Intersection-over-Union (IoU), is one of the most critical evaluation metrics in image semantic segmentation. However, direct optimization of IoU score is very difficult  because the learning objective is neither differentiable nor decomposable. Although some algorithms have been proposed to optimize its surrogates, there is no guarantee provided for the generalization ability. In this paper, we propose a margin calibration method, which can be directly used as a learning objective, for an improved generalization of IoU over the data-distribution, underpinned by a rigid lower bound. This scheme theoretically ensures a better segmentation performance in terms of IoU score. We evaluated the effectiveness of the proposed margin calibration method on seven image datasets, showing substantial improvements in IoU score over other learning objectives using deep segmentation models.
\end{abstract}

\section{Introduction}

Semantic segmentation in images is a fundamental yet challenging problem in computer vision. The task is to build a computational model to accurately assign a class label to every pixel. Semantic segmentation has drawn a broad research interest for many applications such as robotic sensing \cite{ICRA14:ROBOT_SENSING} and auto-navigation \cite{ICCV09:SEG43D}. Recently, the development of deep convolutional neural networks has led to remarkable progress in semantic segmentation due to their powerful feature representation ability to describe the local visual properties. Deep parsing networks are often fine-tuned based on the pre-trained classification networks, e.g., deep residual networks \cite{CVPR16:RESNET}. 

To train a reliable deep learning model for semantic segmentation, the learning objective is one of the most critical ingredients. The most straightforward way is to treat the semantic segmentation as a dense classification task, which examines each pixel in images individually, comparing the class-predictions to the one-hot encoded ground truth. As a surrogate relaxation of the mis-classification rate, cross-entropy becomes the most intuitive loss function in training deep semantic segmentation models. The minimization of cross-entropy is directly related to the maximization of pixel accuracy. In the training process, cross-entropy loss averages over all pixels in images, which is essentially asserting equal learning to each pixel in an image batch. This is problematic in semantic segmentation if the actual classes are imbalanced in the image corpus, as training can be dominated by the most prevalent class, e.g., the small foreground interest regions are submerged by large background areas. Although applying a cost-sensitive re-weighting scheme \cite{MICCAI19:ELL} to alleviate the data imbalance and emphasize the ``important'' pixels, it is unclear how to determine the weights for the best IoU scores \cite{MIA:ODESSEY}, because the pixels of minority class do not necessarily mean they are difficult to be classified (see the ablation study in Section \ref{SEC:ABLATION}). Furthermore, the measure of cross-entropy on the validation set is a poor indicator of the model quality \cite{CVPR18:LOVASZ_SOFTMAX}, as minimizing the pixel-wise loss cannot guarantee a higher IoU score, which is more commonly used in semantic segmentation and can better sketch the contours of interest regions. To address these issues, some recently proposed loss functions have been proposed, e.g., Focal loss \cite{CVPR17:FOCAL} and Lov{\'a}sz-softmax \cite{CVPR18:LOVASZ_SOFTMAX}. The focal loss is an improved cross-entropy loss that tries to handle the class imbalance problem by assigning more weights to hard or easily misclassified examples and down-weight easy examples. The Lov{\'a}sz-softmax is a Lov{\'a}sz surrogate that mimics the IoU, making it consistent with the evaluation metric in semantic segmentation. 

A ``better'' machine learning model should feature a better-generalized performance, i.e., the performance measured on the underlying data distribution, where the unknown instances are sampled from. Clearly, there is a gap between the empirical performance on the training dataset and the generalized performance regarding IoU, i.e., there always exists the IoU differences between training and validation datasets. This gap is commonly called the generalization error. Even though some regularization schemes have been applied to the training of neural networks, their influence on the generalization error of IoU still remains unclear. In this work, we explicitly show how this generalization error is related to label distribution and can be controlled by adding some class-dependent bias terms in the output of the deep semantic segmentation network. These bias terms are connected to the margins among multiple classes, where we are inspired by the idea of margins from the well-known Support Vector Machines (SVMs) \cite{boser1992training}. In \cite{ECCV08:LOCALIZE}, the authors proposed to apply the structured regression to predict the bounding box for object localization. For data-imbalanced learning problems, uneven margins can be applied to well calibrate the importance of specific classes \cite{li2002perceptron,khan2019striking,cao2019learning}. In semantic segmentation, class imbalance widely exists in most image datasets, which hinders the generalization ability of the model, because the IoU score for each class is jointly optimized with others. The power of the ``uneven'' margins inspires us to develop a proper margin calibration scheme for a better generalization ability of semantic segmentation models.

In this paper, we propose a novel distribution-aware margin calibration method, to optimize the IoU in semantic segmentation. The margins across multiple classes are pre-computed based on the label distribution, which can well calibrate the distance between foreground and background classes. Our method has the following three compelling advantages over other learning objectives: (1) it provides a lower bound for data-distribution IoU, which means the model has a guaranteed generalization ability; (2) the margin-offsets can be efficiently computed, which is readily pluggable into deep segmentation models; (3) the proposed learning objective is directly related to IoU scores, i.e., it is consistent with the evaluation metric. Due to the high discriminative power and stability, it is worth using the proposed margin calibration method as a learning objective in the challenging semantic segmentation tasks. We conduct extensive experiments on seven public image datasets, which indicates our method can achieve a considerable improvement compared to other learning objectives.

The rest of the paper is organized as follows. Section \ref{SEC:RELATED_WORK} introduces related work. Section \ref{SEC:METHOD} elaborates the proposed margin calibration method. Experimental results and analysis are presented in Section \ref{SEC:EXP}. Finally, Section \ref{SEC:CONCLUSION} concludes the paper.

\section{Related work} \label{SEC:RELATED_WORK}

\subsection{Deep learning-based semantic segmentation models}

Deep learning-based image segmentation models have achieved significant progress on large-scale benchmark datasets \cite{CVPR17:ADE20K,CVPR16:CITYSCAPES} in recent years. The deep segmentation methods can be generally divided into two streams: the fully-convolutional networks (FCNs) and the encoder-decoder structures. The FCNs \cite{CVPR15:FCN} are mainly designed for general segmentation tasks, such as scene parsing and instance segmentation. Most FCNs are based on a stem-network (e.g., deep residual networks \cite{CVPR16:RESNET}) pre-trained on a large-scale dataset. These classification networks usually stack convolution and down-sampling layers to obtain visual feature maps with rich semantics. The deeper layer features with rich semantics are crucial for accurate classification, but lead to the reduced resolution and in turn spatial information loss. To address this issue, the encoder-decoder structures such as U-Net \cite{MICCAI15:UNET} have been proposed. The encoder maps the original images into low-resolution feature representations, while the decoder mainly restores the spatial information with skip-connections. Another popular method that has been widely used in semantic segmentation is the dilated (atrous) convolution \cite{ARXIV:DILATED}, which can enlarge the receptive field in the feature maps without adding more computation overhead, thus more visual details are preserved. Some methods, such as DeepLab v3+ \cite{ECCV18:DEEPLAB}, just combine the encoder-decoder structure and dilated convolution, to effectively boost the pixel-wise prediction accuracy.

\subsection{Learning objectives for semantic segmentation}

As a dense prediction task, the commonly used cross-entropy is a natural learning objective in training a semantic segmentation model. However, the classification accuracy is inconsistent with the evaluation metric IoU. In recent years, various learning objectives have been proposed specifically for semantic segmentation, and most of them can be used in a plug-and-play way. For example, the distribution-based loss functions (e.g., weighted cross-entropy loss \cite{MICCAI15:UNET} and focal loss \cite{CVPR17:FOCAL}), the region-based loss functions (e.g., IoU loss \cite{ISVC16:IOU}, Dice loss \cite{TMI:DICE} and Tversky loss \cite{MLMIW17:TVERSKY}) and boundary-based loss functions (e.g., Hausdorff distance loss \cite{TMI:HAUSDORFF} and Boundary loss \cite{MIDL19:BOUNDARY}). In medical image segmentation, Ma et al. presented a comprehensive review of 20 general loss functions \cite{MIA:ODESSEY}. These loss functions can also be jointly used in model optimization \cite{ISBI19:FOCAL_TVK}. In deep learning based segmentation methods, the model outputs continuous class probabilities of all pixels, which are indirectly related to IoU scores. To deal with this problem, Maxim et al. proposed to use submodular measures to readily optimize the segmentation model in the continuous setting \cite{CVPR18:LOVASZ_SOFTMAX}. In many real application scenarios, especially scene parsing, the pixel-labels are highly imbalanced, so we prefer to balance the label weights among different classes. By adding down-weights to the well-classified records and assign large weights to misclassified records, focal loss can effectively boost the performance of dense prediction.

On the other hand, design proper surrogates as learning objectives is also applicable to mimic the IoU in semantic segmentation. For example, Nowozin proposed a statistical approximation based on parametric linear programming as a tractable decision making process \cite{CVPR14:DECISIONS}. Ahmed et al. combine the \emph{expected-intersection over expected-union (EIoEU)} with optimizing the \emph{expected-IoU (EIoU)} for a set of candidate solutions \cite{ICCV15:EIOU}. For the deep learning based semantic segmentation, Nagendar et al. proposed to plug a surrogate network into the deep segmentation model for the approximation of IoU \cite{BMVC18:NEUROIOU}, while such a scheme can be also extended to other non-decomposable evaluation metrics, e.g., miss-classification rate (MCR) and Average Precision (AP), in universal machine learning tasks \cite{ARXIV:LEARN_SURROGATE}. 

However, the above methods are mainly to minimize the empirical risk in the model training procedure, without the consideration of the generalization of IoU. In deep learning based segmentation tasks, one may directly use some general methods such as $l_2$-norm, weight-decay, drop-out or extensive data augmentation to improve the generalization ability, but it is unclear how or whether these methods are correlated to the generalization of IoU. As one of the critical objective in design machine learning models, optimizing the generalized performance can be achieved through (1) optimizing the empirical performance approximated by a surrogate loss associated with the performance metric \cite{CVPR18:LOVASZ_SOFTMAX,ARXIV:LEARN_SURROGATE}, e.g. IoU; and (2) controlling the generalized error. In our work, we design a margin calibration scheme with a proper loss function to overcome this difficulty, which provides a better learning objective for semantic segmentation compared to other learning metrics, both theoretically and practically.

\section{Method} \label{SEC:METHOD}

\subsection{Problem setup and notations}

Semantic segmentation in images is essentially a dense classification problem, where a model predicts the one-hot labels to distinguish each foreground class from the background class, using the definition of true positive ($TP$), false positive ($FP$) and false negative ($FN$). The Jaccard index (IoU) is defined as the size of intersection divided by the size of the union of the sample sets:
\begin{equation}
IoU = \frac{TP}{TP+FP+FN}.
\end{equation}
In semantic segmentation, IoU is measured from the pixel-wise classifications, which differs from object localization, where IoU is calculated by the regression of bounding boxes \cite{ECCV08:LOCALIZE}. In this paper, we do not consider the regression case.

A similar metric to IoU is Dice Similarity Coefficient (DSC), which is equivalent to F1-score:
\begin{equation}
DSC = \frac{2\times TP}{2\times TP+FP+FN}.
\end{equation}

However, the above two metrics are count-based measures, whereas the outputs of deep segmentation models are probability values representing the likelihood of the pixels belonging different classes. Therefore, neither IoU score nor DSC can be directly and accurately measured from the output of the network. 

For a multi-class semantic segmentation problem, we formally define an input space $\mathcal{X}\in\mathbb{R}^{w\times h\times c}$ and the target space $\mathcal{Y} =\{1,\ldots, K\}^{w\times h}$, where $w$, $h$, $c$ are the width, height and numbe of channels of an input image, and $K$ is the total number of classes to be segmented. For simplicity, we use $M=w\times h$ to represent the total number of pixels in an image. The function $\theta\in\Theta: \mathcal{X} \mapsto \mathcal{Y}$ is a complex non-linear projection from images to masks (pixel labels). In deep learning-based semantic segmentation methods, $\Theta$ can be a learning framework with trainable parameters. 

Given an image $\mathbf{x}\in\mathcal{X}$ with a corresponding mask $\mathbf{y}\in\mathcal{Y}$, we denote the discrete predicted label for $i$-th pixel is $\hat y_{i}$. Then, given a ground truth $\mathbf{y}$ and a prediction $\mathbf{\hat y}$, the empirical IoU regarding the $k$-th foreground class is:

\begin{equation} \label{miou}
IoU_k = \frac{P_k - P_{k0}}{ P_k + P_{0k}}.
\end{equation}
where $P_{k0}$ denotes the empirical probability that a foreground class $k$ pixel is observed but is predicted as the background class by $\theta$, i.e., 
\begin{equation}
P_{k0}=\frac{1}{M}\sum\limits_{i=1}^{M} \mathbb{I}(y_i=k \wedge \hat{y}_i\neq k),
\end{equation} 
with $\mathbb{I}(\cdot)$ an indicator function. Similarly, $P_{0k}$ denotes the empirical probability that a pixel of the background class is observed but is predicted as the foreground class $k$, i.e., 
\begin{equation}
P_{0k}=\frac{1}{M}\sum\limits_{i=1}^{M} \mathbb{I}(y_i\neq k \wedge \hat{y}_i= k).
\end{equation} 

We use $P_k$ to denote the empirical probability that a class $k$ foreground pixel is observed, i.e., 
\begin{equation}
P_k = \frac{1}{M}\sum\limits_{i=1}^{M}\mathbb{I}(y_i=k).  
\end{equation}

In the evaluation of the segmentation performance, IoU is computed globally over an image dataset, in which the total number of pixels is $N$, where $N\gg M$. From the statistical perspective, we assume that the image samples in the whole dataset are independently and identically distributed (i.i.d) according to some unknown distribution $\mathcal{D}$ over $\mathcal{X}\times \mathcal{Y}$, and let $\mathcal{D}_{\mathcal{Y}}$ denote the projection of $\mathcal{D}$ over $\mathcal{Y}$. Note that we do not assume the pixels in an image are i.i.d. The IoU of $k$-th class over the whole data distribution is:

\begin{equation} \label{emiou}
\mathpzc{IoU}_k = \frac{\mathpzc{P}_k - \mathpzc{P}_{k0}}{ \mathpzc{P}_k + \mathpzc{P}_{0k}}.
\end{equation}

Note that we use $P$ and $IoU$ of the normal font to represent the empirical probability and IoU on the finite image dataset, while use  $\mathpzc{P}$ and $\mathpzc{IoU}$ of the calligraphic font to represent the probability and IoU over the whole data distribution.

When there are $K$ classes presented, the empirical mean IoU (mIoU) is defined as $mIoU = \frac{1}{K} \sum\limits_{k=1}^{K} IoU_k$, and similarly, the mIoU over the data distribution $\mathcal{D}$ is defined as $\mathpzc{mIoU} = \frac{1}{K} \sum\limits_{k=1}^{K} \mathpzc{IoU}_k$.

Ideally, a function $\theta$ should produce a high $\mathpzc{mIoU}$ over the data distribution to ensure the stable performance of $\theta$ on any data sampled from $\mathcal{D}$. Unfortunately, the data distribution $\mathcal{D}$ is usually fixed but unknown. Thus, we can only optimize the empirical mIoU, so that with a high probability it can lead to a high $\mathpzc{mIoU}$ over $\mathcal{D}$. The problem here is that how $mIoU$ and $\mathpzc{mIoU}$ are close to each other. Next, we present our method to minimize the error bound between $mIoU$ and $\mathpzc{mIoU}$, which can theoretically support the proposed margin calibration method.

\subsection{Method overview}

In semantic segmentation tasks, the label imbalance is an inherent issue for dense prediction, so equally treating all pixel labels in the model training may lead to the biased IoU scores towards the majority classes. Since in deep semantic segmentation models, the one-hot pixel labels of multiple classes are simultaneously optimized, the minority class may be still under-fitted when the majority class is already over-fitted. An intuitive approach is to set different weights to the loss function based on the number of pixels of each objective class, e.g., weighted cross-entropy. However, it is unclear if the weights of the loss functions based on the number of total pixels of the objective classes are optimal, because the ``hardness'' of segmenting the minority classes is not directly related to the number of training pixels. In our own experience, we found that weighted cross-entropy barely improves the model performance in terms of IoU.

\begin{figure}[t]\center
	\includegraphics[width=0.4\textwidth]{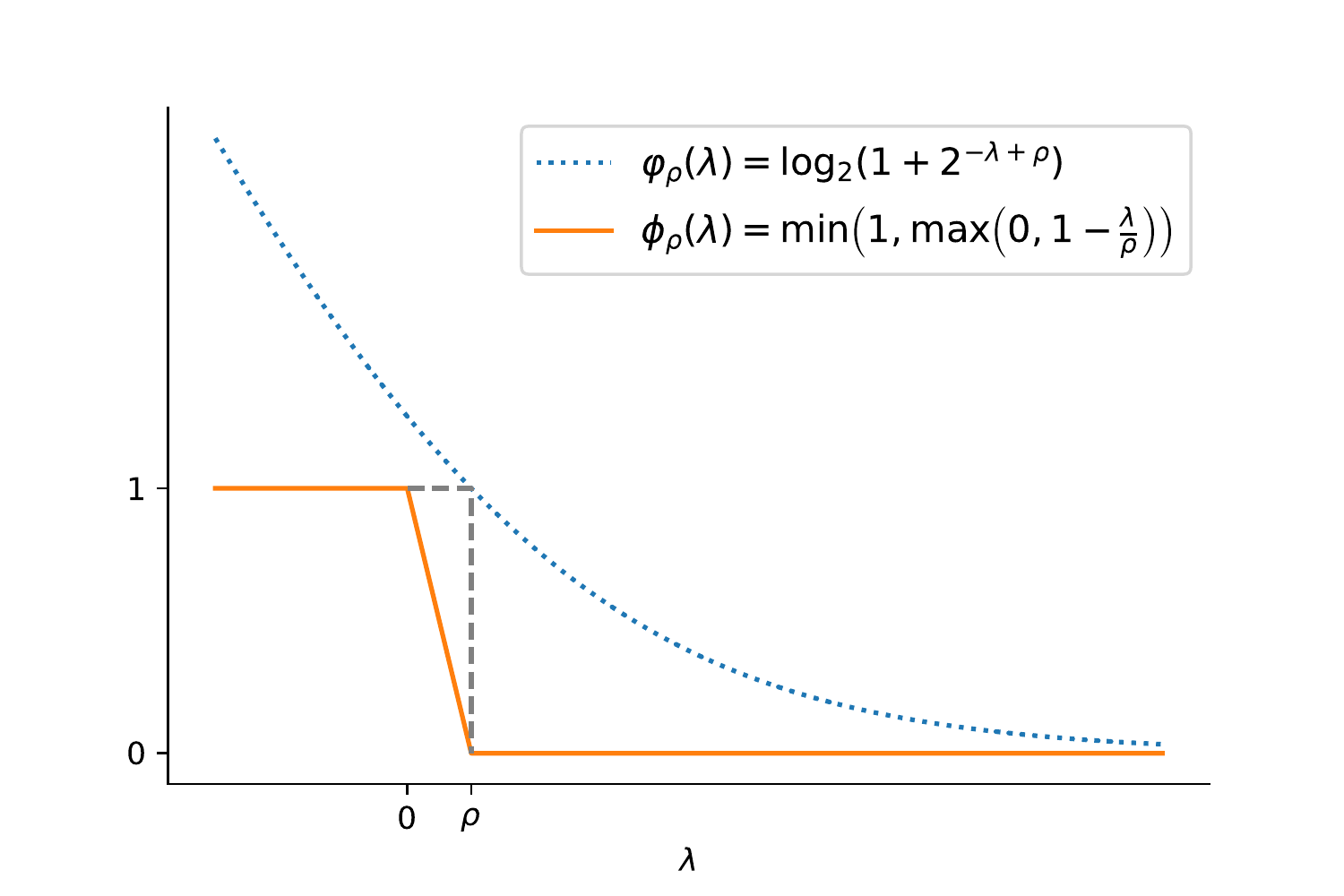}
	\caption{The $\rho$-calibrated log-loss (blue dotted line) and $\rho$-margin loss (orange solid line) functions. The $\rho$-margin loss is a upper bound for 0-1 loss. For the $\rho$-calibrated log-loss, $\varphi_\rho(\rho)=1$ and it upper bounds the $\rho$-margin loss.} \label{lossfunplot}
\end{figure}

In our approach, we instead set different margins for the pixel classes, which differs from the weight loss functions. Specifically, we would derive an optimal margin setting for a small error bound between $mIoU$ and $\mathpzc{mIoU}$. Denote the output score of $i$-th pixel in the image dataset regarding the $k$-th foreground class by $s_{ik}$. Here we define the {\bf margin} for the $i$-th pixel with regard to class $k$ in the whole image set as:
\begin{equation}\label{calmargin}
\lambda_{ik}=s_{ik}- \max_{j\neq k} s_{ij}.
\end{equation}
If the $i$-th pixel belongs to $k$-th foreground class, it is preferable to have a large positive value of $\lambda_{ik}$. Otherwise, we expect it to be a negative value. We then combine the margin $\lambda_{ik}$  with a $\rho$-margin loss function $\phi_\rho(\cdot)$ defined in \cite[Definition 5.5]{mohri2018foundations}, to build the relationship between IoU score and the margin $\lambda_{ik}$. The $\rho$-margin loss is defined as: 
\begin{equation}\label{rho_margin}
\phi_{\rho}(\lambda)=\min \left(1, \max \left(0,1-\frac{\lambda}{\rho}\right)\right),
\end{equation}
which encourages the margin $\lambda$ to be larger than $\rho$ and provides an upper bound for 0-1 loss, as is illustrated in Fig. \ref{lossfunplot}. We call the parameter $\rho$ \textbf{margin-offset}. We can then bound the empirical probabilities $P_{k0}$ and $P_{0k}$ in Eq.(\ref{miou}) as: 
\begin{equation}\label{rhomloss}
\begin{aligned}
P_{k0}(\theta) & <\frac{1}{N}\sum\limits_{i\in Y_k} \phi_{\rho_{k0}}(\lambda_{ik}) = \ell_{k0}(\theta,\rho_{k0}), \\ 
P_{0k}(\theta) & <\frac{1}{N}\sum\limits_{i\in Y\setminus Y_k} \phi_{\rho_{0k}}(-\lambda_{ik}) = \ell_{0k}(\theta,\rho_{0k}),
\end{aligned}
\end{equation} 
where we use $Y_k$ and $Y\setminus Y_k$ to denote the index set of the foreground pixels of $k$-th class and background pixels, respectively. $\rho_{0k}$ and $\rho_{k0}$ are pre-defined margin-offsets. Then, we can give a lower bound for Eq.(\ref{miou}) as:
\begin{equation}\label{apploss}
\overline{IoU}_k = \frac{P_{k} - \ell_{k0}(\theta,\rho_{k0})}{ P_{k} + \ell_{0k}(\theta,\rho_{0k})},
\end{equation} 
and the corresponding lower bound for $mIoU$ is:
\begin{equation}
\overline{mIoU} = \frac{1}{K}\sum\limits_{k=1}^{K}\overline{IoU}_k.
\end{equation}

\subsection{Theoretical motivation}

We can derive a generalization error bound regarding IoU with the margin-offsets $\rho_{k0}$ and $\rho_{0k}$, based on the following theorem:
\begin{theorem}\label{miongen}
For any function $\theta\in\Theta$, define $\mu_k=\frac{\rho_{k0}}{\rho_{0k}}$ and $\mathcal{F}=C(\Theta)+\sigma(\frac{1}{\eta})$. $C(\Theta)$ is some proper complexity measure of the hypothesis class $\Theta$ and $\sigma(\frac{1}{\eta}) \triangleq \frac{\rho_{\max}}{4K} \sqrt{2M\log \frac{2K}{\eta}}$ is typically a low-order term in $\frac{1}{\eta}$ with $\rho_{\max} = \max \{\rho_{k0},\rho_{0k}\}_{k=1}^K$. Given a training dataset of $N$ image pixels including $N_k$ pixels of class $k$, with each image consisting of $M$ pixels, then for any $\eta>0$, with probability at least $1-\eta$,
\begin{equation}\label{tobepf}
	\mathpzc{mIoU} \geq \overline{mIoU}-\epsilon,
\end{equation}
where 
\begin{equation}\label{epsilon}
\epsilon = \frac{1}{K}\sum\limits_{k=1}^{K} \frac{\sqrt{N-N_k} + \frac{\sqrt{N_k}}{\mu_k}}{\frac{N_k}{4K\mathcal{F}}\rho_{0k}-\sqrt{N-N_k}}.
\end{equation}
\end{theorem}
Note that this theorem involves a complexity measure $\mathcal{F}=C(\Theta)+\sigma(\frac{1}{\eta})$, where $C(\Theta)$ is derived from the Rademacher complexity. The Rademacher complexity typically scales in $\sqrt{\frac{C(\Theta)}{N_k}}$ \cite{mohri2018foundations}. Such a scale has been used in related works (see \cite{cao2019learning,neyshabur2018role} and the references therein) to imply a connection between Rademacher complexity and number of pixels $N_k$. See the proof of the theorem in the appendix. This theorem enables us to maximize the $\mathpzc{mIoU}$ on the data distribution by maximizing a lower bound $\overline{mIoU}$ for the empirical IoU on a training dataset with a high probability. Meanwhile, we would prefer a small error bound $\epsilon$ so that the lower bound $\overline{mIoU}$ on the empirical IoU could be a reliable estimation for $\mathpzc{mIoU}$. This scheme guarantees the performance of associated function $\theta$ on the unseen image data.

\textbf{Remark.} At first glance, the relationship between $N$, $N_k$ and $\epsilon$ seems complicated. However, $\epsilon$ decreases if we increase $N$ and $N_k$ proportionally, as can be inferred from Eq. (\ref{epsilon}), so decreasing $\epsilon$ would need more these pixels accordingly. Theorem \ref{miongen} indicates that a smaller $\epsilon$ requires more foreground class pixels, and a simple fit function for a smaller $C(\Theta)$. Another important factor is that we can adjust the margin-offset $\rho_{0k}$ to minimize the error bound $\epsilon$. Note that increasing $\rho_{0k}$ also increases the $C(\Theta)$ implicitly, because a larger margin-offset may require more complex hypothesis class $\Theta$. Otherwise, $\overline{mIoU}$ may decrease due to the under-fitting. Besides, the direct calculation of the optimal margin-offsets in Theorem \ref{miongen} is difficult because it is related the complexity measure $C(\Theta)$, which is measured by the structure of deep neural networks. Nevertheless, we can give the optimal $\rho_{0k}$ that is irrelevant to $C(\Theta)$, by the following corollary:
\begin{corollary}\label{proprho}
Assume $\sum\limits_{k=1}^{K}\rho_{0k} = \text{some constant}.$ Let $\mu_k = \frac{P_{k}\sqrt{N_k}}{\upsilon(N-N_k)-P_{k}\sqrt{N-N_k}}$ with $\upsilon$ ($\upsilon>0$) being a hyper-parameter. Then the minimum of the error bound $\epsilon$ in Theorem \ref{miongen} is attained given the following condition:
\begin{equation}
\frac{\rho_{0i}}{\rho_{0j}} = \frac{N_j}{N_i} \frac{\sqrt{N-N_i}}{\sqrt{N-N_j}}, \quad \frac{\rho_{k0}}{\rho_{0k}}=\mu_k, \quad \forall i,j,k\in[1,K].
\end{equation} 
\end{corollary}

See the proof of the corollary in the appendix. Corollary \ref{proprho} provides a theoretical guarantee for setting the margin-offsets towards a smaller error bound $\epsilon$. The margin-offset $\rho_{0k}$ is proportional to $\frac{\sqrt{N-N_k}}{N_k}$, which indicates a larger margin is required for $k$-th class, with comparably fewer pixels. We introduce a hyper-parameter $\tau$ ($\tau>0$) to scale the margin-offsets, which can be tuned on the validation dataset. Note that another margin $\rho_{k0}=\mu_k\rho_{0k}$ is usually small compared to $\rho_{0k}$ in practice with a well-tuned $\upsilon$, so we can mainly focus on $\rho_{0k}$ here. A proper setting of $\tau$ and $\upsilon$ can provide a balance between $\epsilon$ and $\overline{mIoU}$ for the maximization of $\mathpzc{mIoU}$. Empirically, just setting $\tau=10$ and $\upsilon=1$ can obtain a satisfactory result.

\subsection{A practical implementation with $\rho$-calibrated log-loss function}

Based on the above statistical analysis of label distribution, the learning objective needs to compute the margin-offsets from the whole pixel label set before optimizing the network. Given a pixel label set with $K$ classes, the computation of margin-offsets is summarized in Algorithm \ref{algomar1}.

\begin{algorithm}
	\caption{Margin-offsets calculation} 
	\label{algomar1}
    \textbf{Input}: Labels for all pixels of size $N$. Number of pixels in each class given by $N_1,...,N_K$. Hyper-parameter $\upsilon, \tau > 0$. \\
    \textbf{Output}: The margin-offsets $\rho_{0k}$, $\rho_{k0}$, $\forall k \in [1,K]$ 
	\begin{algorithmic}[1]
		\For {class $k=1,2,\ldots,K$}
		    \State $P_k = \frac{N_k}{N}$;
		    \State $\mu_k = \frac{P_{k}\sqrt{N_k}}{\upsilon(N-N_k)-P_{k}\sqrt{N-N_k}}$;
			\State $\rho_{0k} = \tau \times \frac{\sqrt{N-N_k}}{N_k}$;
			\State $\rho_{k0} = \mu_k\rho_{0k}$.
		\EndFor 
	\end{algorithmic} 
\end{algorithm}

The learning objective of semantic segmentation is to maximize $\mathpzc{mIoU}$ for the best performance. Ideally, we should maximize its lower bound $\overline{mIoU}$ with a small error bound $\epsilon$, because the margin-offset can provide a guarantee for its generalization. However, in the training of deep neural networks, the direct optimization of $\overline{mIoU}$ is impractical because the network is trained in a mini-batch manner. Unlike other decomposable evaluation metrics, such as classification accuracy, where the expectation of the metric on a mini-batch sample is equivalent to the metric on the whole dataset, the expectation of the mini-batch mIoU is an estimation to the overall mIoU on the whole dataset, i.e., the empirical IoU on the training dataset may be sub-optimal.  
 
For a practical implementation, we instead minimize the sum of $\rho$-margin losses $l_{k0}$ and $l_{0k}$ involved $\overline{mIoU}$, with the optimal margin-offset. So for a mini-batch images, the loss $L(\theta)$ is calculated by:

\begin{align}\label{mgloss}
L(\theta) &= \sum\limits_{k=1}^{K}(\ell_{k0}(\theta,\rho_{k0})+\ell_{0k}(\theta,\rho_{0k})) \nonumber\\
&= \frac{1}{N_s}\sum\limits_{k=1}^{K}\left(\sum\limits_{i\in Y_k} \phi_{\rho_{k0}}(\lambda_{ik})+\sum\limits_{i\in Y\setminus Y_k} \phi_{\rho_{0k}}(-\lambda_{ik})\right),
\end{align}
with $\lambda_{ik}$ defined in Eq.(\ref{calmargin}), and $N_s$ is the number of pixels in a mini-batch. In the forward pass of training, the network $\theta$ outputs a batch of pixel-wise scores, in which the $i$-th pixel with regard to class $k$ is $s_{ik}$. Then $s_{ik}$ is used to calculate $\lambda_{ik}$ by Eq.(\ref{calmargin}).

In practice, the non-smoothness of $\rho$-margin loss function may bring instability in the optimization. As is shown in Fig. \ref{lossfunplot}, the gradient regarding the $\rho$-margin loss can be prohibitively large when $\rho$ is very small, while the gradients outside the interval $(0,\rho)$ is zero. Thus, we substitute the $\rho$-margin loss $\phi_\rho(\lambda)$ used in Eq.(\ref{mgloss}) with $\rho$-calibrated log-loss $\varphi_\rho(\lambda)=\log_2(1+2^{-\lambda+\rho})$. The relationship between the $\rho$-margin loss $\phi_\rho(\lambda)$ and the $\rho$-calibrated log-loss $\varphi_\rho(\lambda)$ is illustrated in Fig. \ref{lossfunplot}. Now we apply the margin-offsets to get a biased score $\bar{s}_{ik}$ for $\rho$-margin loss. The computation of $\bar{s}_{ik}$ is described in Algorithm \ref{algomar2}.

\begin{algorithm}
	\caption{$\rho$-margin calibration} 
	\label{algomar2}
    \textbf{Input}: Prediction scores $\mathbf{s}=\{s_{ik}\}$, ground truth $y_i$, the margin-offset $\rho_{0k}$, $\rho_{k0}$, $\forall k \in [1,K]$ and $\forall i \in [1,N_s]$. \\
    \textbf{Output}: $\rho$-margin calibrated prediction scores $\bar{\mathbf{s}} =\{\bar{s}_{ik}\}$.
	\begin{algorithmic}[1]
	\For {pixel $i=1,\ldots, N_s$}
		\For {class $k=1,2,\ldots,K$}
           \State $\lambda_{ik}=s_{ik}- \max\limits_{j\neq k}s_{ij}$,
           \State $\bar{s}_{ik}=
				\begin{cases}
				\lambda_{ik} - \rho_{k0} & \text{if} \quad y_i=k, \\
				\lambda_{ik} + \rho_{0k} & \text{else}.  \nonumber
				\end{cases}$		           
	    \EndFor	
	\EndFor	
	\end{algorithmic} 
\end{algorithm}

For the use of the $\rho$-calibrated log-loss $\varphi_\rho(\lambda)$, we first calibrate the output $\{s_{ik}\}$ via Algorithm \ref{algomar2}, then the $\rho$-calibrated log-loss bounds the $\rho$-margin loss from above and leads to:
\begin{equation}
\ell_{k0}(\theta,\rho_{k0})\!<\!\frac{1}{N_s}\sum\limits_{i\in Y_k} \log_2(1\!+\!2^{-\bar{s}_{ik}})=\underline{\ell_{k0}}(\theta,\rho_{k0}),
\end{equation}
and
\begin{equation}
\ell_{0k}(\theta,\rho_{0k})\!<\!\frac{1}{N_s}\sum\limits_{i\in Y\setminus Y_k} \log_2(1\!+\!2^{\bar{s}_{ik}})=\underline{\ell_{0k}}(\theta,\rho_{0k}).
\end{equation}

Based on the above two inequalities, we simply use $\underline{\ell_{k0}}(\theta,\rho_{k0})$ and $\underline{\ell_{0k}}(\theta,\rho_{0k})$ to replace $\ell_{k0}(\theta,\rho_{k0})$ and $\ell_{0k}(\theta,\rho_{0k})$ in Eq.(\ref{mgloss}) as the final learning objective.

\subsection{Complexity analysis}

Given the output scores $\{s_{ik}\}^{N_s\times K}$ of $N_s$ pixels in an image batch, with the parallel computation provided by GPUs, calculating the margin $\lambda_{ik}$ needs $O(N_s)$ time and $O(N_s K)$ space, and the subsequent calibrated log-loss incurs $O(N_sK)$ time complexity. So compared to the cross-entropy loss, the calibration method requires extra $O(N_s+N_sK)$ time and $O(N_s K)$ space complexities overhead in computing the calibrated log-loss.

\subsection{Discussions}

How to optimize non-decomposable loss like IoU is an open problem. This problem becomes far more challenging in the mini-batch training setting because in this case, we optimize the mini-batch IoU, which is an estimation of the overall IoU on the whole dataset. In our method, we mainly deal with a ratio distribution (both denominator and numerator of IoU are random variables regarding data distribution), where the central limit theorem can not be applied. As such, currently, no method can deal with this mini-batch setting accurately including lov{\'a}sz-softmax, which claims to be a surrogate for optimizing IoU. We also compromise on the learning objectives to optimize a related $\rho$-margin calibrated log-loss, which is independent of the margin calibration process. This makes our IoU on the training set a more reliable indicator for the IoU over the underlying distribution than other methods.

In the deep learning based semantic segmentation settings, directly applying margin calibration incurs additional space and time complexities. Consequently, the computation may be slower and more memory-consuming. To alleviate this, the network parameters can be initialized via training with cross-entropy, the most efficient but not the ``perfect'' learning objective, then fine-tuned with margin-calibration.

\section{Experiment} \label{SEC:EXP}

In this section, we use the deep segmentation models to conduct the experiments on five publicly available datasets. Unlike the proposal of deep learning architectures that aims to achieve the best segmentation performance in some recent works \cite{CVPR20:SR,NIPS20:RANET,TIP:CEMD}, our contribution is mainly on the design of a novel learning method for better IoU scores when the learning framework is fixed. Given a deep segmentation model, we mainly compare the final performance when applying commonly used learning objectives and our margin calibration method. 

\subsection{Datasets}

We conducted the experiment of semantic segmentation on seven datasets: Robotic Instrument \cite{ARXIV:ROBOTIC}, COCO-Stuff 10K \cite{CVPR18:COCO}, PASCAL VOC2012 \cite{IJCV:VOC},  MIT SceneParse150 \cite{IJCV:ADE20K}, Cityscapes \cite{CVPR16:CITYSCAPES}, BDD100K \cite{CVPR20:BDD100K} and Mapillary Vistas \cite{CVPR17:MAPILLARY}.

The Robotic Instrument dataset provides 8 robotic surgical videos, in which 225 frames are sampled from each video. In the frames, each part is manually annotated by a trained team. Here we conduct the instrument part segmentation as an ablation study, in which we aim to correctly segment each articulating part of the instrument.

The COCO-Stuff 10K contains 9,000 images for training and 1,000 images for validation (testing). Following \cite{CVPR18:CCFGMA}, we evaluate the IoU performance on 171 categories (80 objects and 91 stuff) to each pixel.

The PASCAL VOC 2012 semantic benchmark contains 20 foreground object classes and one background class. The original dataset has 1,464 and 1,449 images for training and validation, respectively. To augment the training dataset, we also use extra annotations provided by \cite{ICCV11:VOC}. The model is not pre-trained with MS COCO dataset.

The MIT SceneParse150 dataset is built based on ADE20K \cite{CVPR17:ADE20K} as a scene parsing benchmark. It contains more than 20K scene images, annotated by 150 classes of dense labels. Here we use 2,000 validation images for qualitative evaluation. 

The Cityscapes, BDD100K and Mapillary Vistas are three different street-view datasets. The data in Cityscapes were taken from 50 European cities, which provides fine-grained pixel-level annotations of 19 classes including buildings, pedestrians, bicycles, cars, etc. The training/validation/testing splits are with 2,975, 500 and 1,525 images, respectively. It also has 20,000 coarsely-labelled images, which can be used to pre-train the segmentation model.

Different from the Cityscapes dataset comes from Germany, the images of BDD100K are mainly from the US cities, and there is a dramatic domain shift between the two datasets for semantic segmentation models, although their labels are the same. 

The Mapillary Vistas dataset contains 25,000 high-resolution images annotated into 66 fine-grained object categories, featuring locations from all around the world, and taken from a diverse source of image capturing devices.

\subsection{Settings}

We implemented the segmentation model based on PyTorch\footnote{See our PyTorch implementation at \url{https://github.com/yutao1008/margin_calibration} for more details.}. For the experiment on the Robotic Instrument dataset, we applied the recently proposed COPLE-Net \cite{TMI:COPLENET}, a variant of U-Net \cite{MICCAI15:UNET} for medical image segmentation. COPLE-Net has much fewer trainable parameters compared to FCN, thus it is quite suitable for the segmentation tasks in a simple yet fixed application scenario. On the rest four datasets for general semantic segmentation tasks, we used DeepLab v3+ \cite{TPAMI:DEEPLAB} backend on SEResNeXt-50 \cite{CVPR18:SE}, with the output stride 8. We employed the AdamW optimizer \cite{ICLR19:ADAMW} with the initial learning rate $10^{-4}$ in the training process. We used the mixed precision and gradient checkpoint, which allow to set a larger mini-batch size and can effectively save the GPU memory usage without hurting the batch normalization layers. Our experiments were conducted on a server equipped with a single NVIDIA Tesla V100 GPU card. 

\subsection{Ablation study on Robotic Instrument dataset}\label{SEC:ABLATION}

We sequentially split the Robotic Instrument dataset into 1,200, 200 and 400 images according to the frame index for training, validation and testing, respectively. For this task, we aim to accurately label four instrument parts, including shaft, wrist, claspers and probe. The pixel ratios in the 4 parts are 4.9\%, 1.4\%, 1.6\% and 0.8\%, respectively, while the background class occupies 91.2\% of the total pixels, making the label distribution extremely imbalanced. 

We used categorical cross-entropy as the baseline, as semantic segmentation can be treated as a dense prediction for each image pixel. In addition, we tested several recently proposed methods, including focal loss \cite{CVPR17:FOCAL} (with the scale factor 0.4), lov{\'a}sz-softmax \cite{CVPR18:LOVASZ_SOFTMAX}, generalized dice loss \cite{GDICEV2} and Tversky loss \cite{TVERSKY}. All these methods were used as independent learning objectives in the medical segmentation tasks, and the segmentation networks were all trained from the random state.

\begin{figure*}[t]\centering\small
\begin{minipage}{0.3\textwidth}\centering
	\includegraphics[width=1\textwidth]{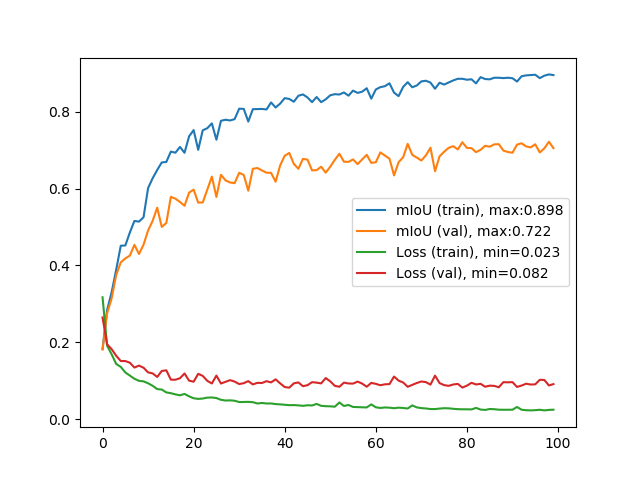}
	(a) Cross entropy 
\end{minipage}
\begin{minipage}{0.3\textwidth}\centering
	\includegraphics[width=1\textwidth]{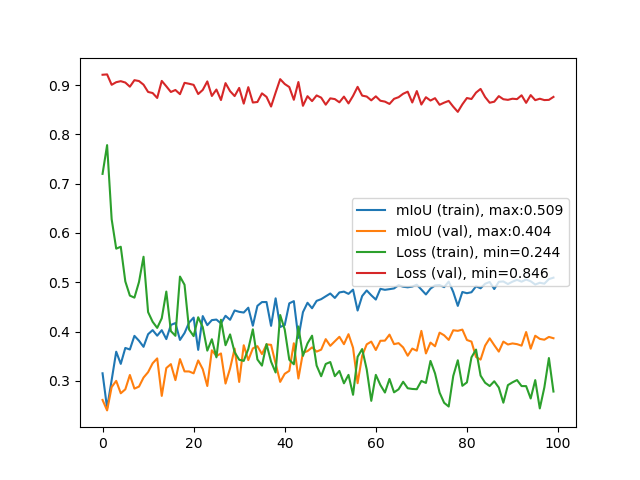}
	(b) Dice loss
\end{minipage}
\begin{minipage}{0.3\textwidth}\centering
	\includegraphics[width=1\textwidth]{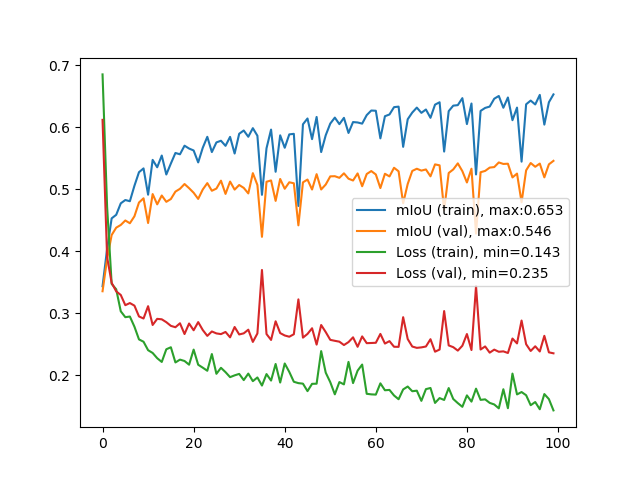}
	(c) Tversky loss
\end{minipage}
\begin{minipage}{0.3\textwidth}\centering	
	\includegraphics[width=1\textwidth]{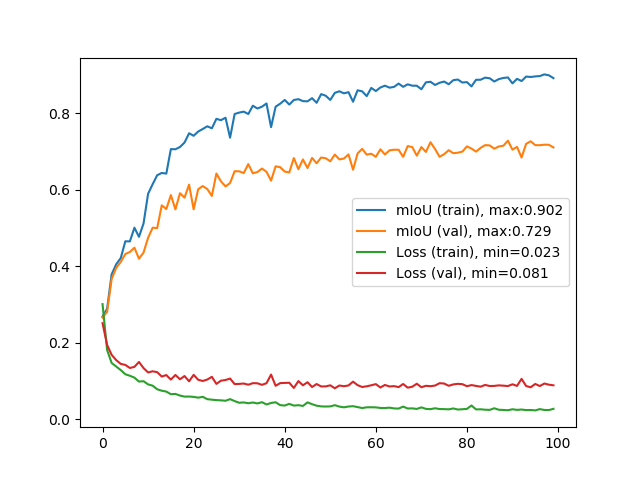} 
	(d) Focal loss   	
\end{minipage}
\begin{minipage}{0.3\textwidth}\centering
	\includegraphics[width=1\textwidth]{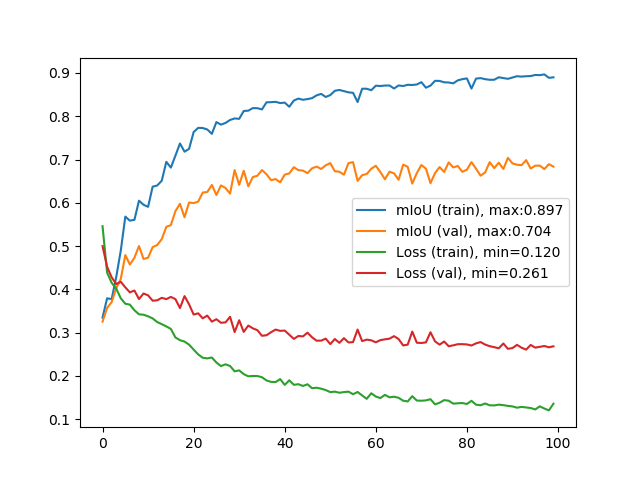}
	(e) Lov{\'a}sz-softmax 
\end{minipage}
\begin{minipage}{0.3\textwidth}\centering
	\includegraphics[width=1\textwidth]{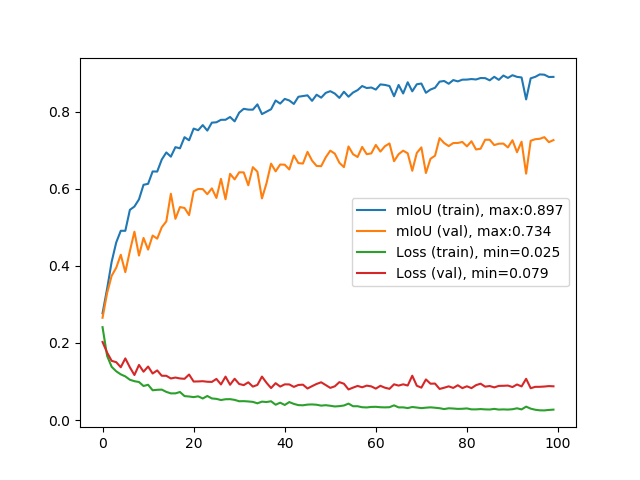}
	(f) Margin calibration
\end{minipage}

\caption{Training mIoU and loss curves on Robotic Instrument dataset.}
\label{FIG:CURVES}
\end{figure*}

We recorded all the intermediate results during the training process, as is shown in Fig. \ref{FIG:CURVES}. By observing these figures, we can see directly using the two count-based loss functions, Dice loss and Tversky loss, the segmentation network fails to converge, as they are not differentiable to pixel-wise categories and can only work in conjunction with distribution-based loss functions. The rest four methods show similar mIoU and loss curves. However, although lov{\'a}sz-softmax can utilize the sub-modular property to minimize the Jaccard loss, it is more likely to over-fit. The proposed margin calibration method can generally act as a plug-and-play learning objective like cross-entropy and focal loss in training a semantic segmentation network. 

In the model inference, we did not use horizontal flipping, multi-scale prediction or CRF post-processing to augment the segmentation performance. The quantitative results are shown in Table \ref{TB:OVERALL_ROBOTIC} and \ref{TB:ROBOTIC_PERCLASS}. The two evaluation metrics, pixel accuracy and IoU score, although have a very high correlation in terms of the absolute values, the best one single metric cannot guarantee the other. For example, simply using cross-entropy achieve a better pixel accuracy compared to the focal loss and lov{\'a}sz-softmax, but its IoU score is the worst. In semantic segmentation, the IoU score is usually a better evaluation to quantify the percent overlap between the pixel-label output and target mask. Using Dice loss as the learning objective, the model fails to segment {\em shaft} and {\em claspers}. Similarly, using Tversky loss cannot segment {\em probe}. Compared with cross-entropy, focal loss and Lov{\'a}sz-softmax, the proposed margin calibration method obtains the best pixel accuracy and IoU scores on this dataset. Specifically, as a single learning objective, margin calibration outperforms the second-best, with 3.0\% performance gain in terms of the mIoU score. Although margin calibration is not specifically designed to optimize the pixel accuracy, it can also benefit from the generalization ability. 

\begin{figure*}[h]\centering\small
\begin{minipage}{0.23\textwidth}\centering
	\includegraphics[width=1\textwidth]{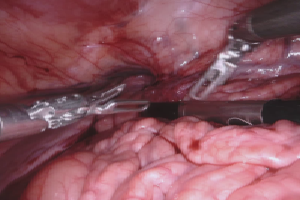}
	(a) Image
\end{minipage}
\begin{minipage}{0.23\textwidth}\centering
	\includegraphics[width=1\textwidth]{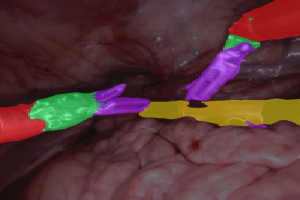}
	(b) Cross-entropy
\end{minipage}
\begin{minipage}{0.23\textwidth}\centering
	\includegraphics[width=1\textwidth]{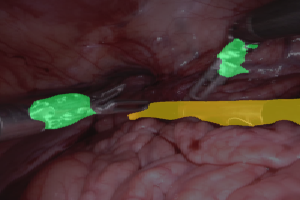}
	(c) Dice loss
\end{minipage}
\begin{minipage}{0.23\textwidth}\centering
	\includegraphics[width=1\textwidth]{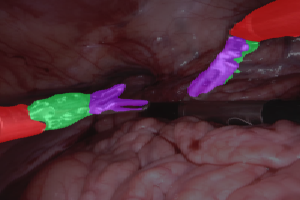}
	(d) Tversky loss 
\end{minipage}
\begin{minipage}{0.23\textwidth}\centering
	\includegraphics[width=1\textwidth]{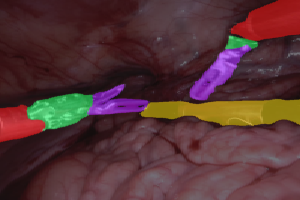}
	(e) Focal loss
\end{minipage}
\begin{minipage}{0.23\textwidth}\centering
	\includegraphics[width=1\textwidth]{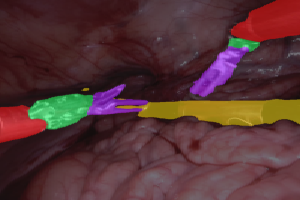}
	(f) Lov{\'a}sz-softmax 
\end{minipage}
\begin{minipage}{0.23\textwidth}\centering
	\includegraphics[width=1\textwidth]{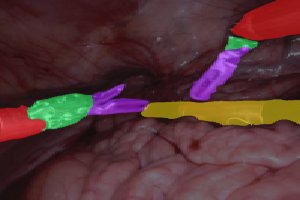}
	(g) Margin calibration
\end{minipage}
\begin{minipage}{0.23\textwidth}\centering
	\includegraphics[width=1\textwidth]{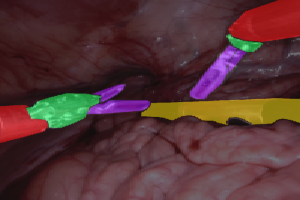}
	(h) Ground truth
\end{minipage}
\caption{Segmentation examples on the Robotic Instrument test set.}
\label{FIG:ROBOTIC}
\end{figure*}

We illustrate the segmentation examples in Fig. \ref{FIG:ROBOTIC}. By observing the results, we can see that applying the proposed margin calibration method can effectively reduce the false positives, forming more smooth contours and obtaining more accurate results.

\begin{table}[t]
\centering \small
\caption{The overall segmentation performance on the Robotic Instrument test set.}
\label{TB:OVERALL_ROBOTIC}
\begin{tabular}{|c||cc|}
\hline
Method & Pixel Acc. & mIoU \\
\hline
Cross-entropy &81.1  &66.2  \\
Dice loss & 38.3 & 27.3\\
Tversky loss & 59.7 &43.8\\
Focal loss   &80.0  &69.5  \\
Lov{\'a}sz-softmax &80.1  &68.9 \\
\hline
Margin calibration &{\bf 81.5}  &{\bf 72.5}   \\
\hline
\end{tabular}
\end{table}

\begin{table}[t]
\centering \small
\caption{Per-class IoU on the Robotic instrument test set.}
\label{TB:ROBOTIC_PERCLASS}
\begin{tabular}{|c||cccc|}
\hline
Method & Shaft & Wrist & Claspers & Probe \\
\hline
Cross-entropy & 81.1 & 55.3 & 55.5 & 72.9 \\
Dice loss & 0.0 &58.1 &0.0 &72.6\\
Tversky loss &83.5 &60.1 &52.9 &0.0\\
Focal loss   & 86.5 & 62.9 & 56.5 & 72.0 \\
Lov{\'a}sz-softmax & 86.3 & 64.4 & 55.5 & 69.3 \\
\hline
Margin calibration & {\bf 88.2} & {\bf 67.1} & {\bf 61.1} & {\bf 73.4} \\
\hline
\end{tabular}
\end{table}

\subsection{Results on COCO-Stuff 10K and PASCAL VOC2012 datasets}

In the training of the segmentation model, we re-scaled the shorter image size to 400 then randomly cropped it to $384\times384$. In the inference, we used the single-scale inference without flipping or any other augmentations. The comparisons on the two validation sets with the three baseline methods are reported in Table \ref{TB:COCO} and Table \ref{TB:VOC}, respectively. Results show that when fixing the deep neural network, both focal loss and Lov{\'a}sz-softmax and improve the mIoU scores. By pre-computing the margin-offsets and applying the proposed margin calibration method, a single segmentation model can achieve a further 0.4\% of the mIoU score on the COCO-Stuff 10K, while on the PASCAL VOC2012 dataset, the mIoU score with margin calibration is only 0.1\% higher than the second-best.

By observing the two datasets, we found that they have different properties regarding pixel-label annotations. The dense prediction on the COCO-Stuff 10K dataset aims to auto-label all pixel classes, while on the PASCAL VOC 2012 the task is to distinguish one or two foreground objects from the unique background class. Also, the label imbalance in COCO-Stuff 10K is much more imminent than PASCAL VOC 2012, leading to the significant difference in margin calibrations. On the PASCAL VOC 2012 dataset, the background class {\em void} occupies 80\% of the total pixels in the image corpus, while each foreground class has a very similar number of pixels (around 1\%). So applying the $\rho$-margin calibration in Algorithm \ref{algomar2}, the learning objective is more like a scaled log-loss.

\begin{table}[t]
\centering \small
\caption{The segmentation performance on the COCO-Stuff 10K validation set.}
\label{TB:COCO}
\begin{tabular}{|c|c|}
\hline
{\bf Method}	 & mIoU   \\
\hline
Cross-entropy 	&  34.1 \\
Focal loss & 34.9 \\
Lov{\'a}sz-softmax & 35.1  \\
Margin calibration & {\bf 35.5} \\
\hline
\end{tabular}
\end{table} 

\begin{table}[t]
\centering \small
\caption{The segmentation performance on the PASCAL VOC2012 validation set.}
\label{TB:VOC}
\begin{tabular}{|c|c|}
\hline
{\bf Method}	 & mIoU   \\
\hline
Cross-entropy 	& 78.2  \\
Focal loss &  78.3 \\
Lov{\'a}sz-softmax & 78.5  \\
Margin calibration & {\bf 78.6} \\
\hline
\end{tabular}
\end{table}

\subsection{Results on MIT SceneParse150 dataset}

We experimented on the large-scale MIT SceneParse150 dataset to verify the effectiveness of the margin calibration method. Unlike the multi-task learning framework such as \cite{ECCV18:UPERNET} to use multiple supervised information for the best segmentation performance, we only use the 150 scene labels and compare different single learning objectives in training the semantic segmentation model. The images were rescaled to $384\times 384$. On this dataset, we first used cross-entropy as the default learning objective to train the DeepLab v3+ model, then fine-tuned the network with focal loss, Lov{\'a}sz-softmax and the proposed margin calibration independently, to see the performance improvement of mIoU scores. The ground truth of the testing set has not been released, so we use the 2,000 validation images for qualitative evaluation. In the model inference, we adopted horizontal flipping and multi-scale prediction to augment the segmentation performance.

\begin{figure*}[t]\centering
\begin{minipage}{0.2\textwidth}
	\includegraphics[width=1\textwidth]{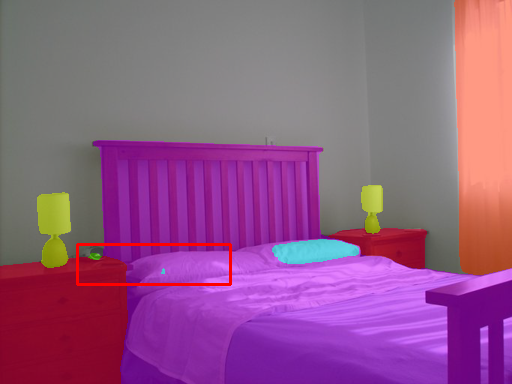}
\end{minipage}
\begin{minipage}{0.2\textwidth}
	\includegraphics[width=1\textwidth]{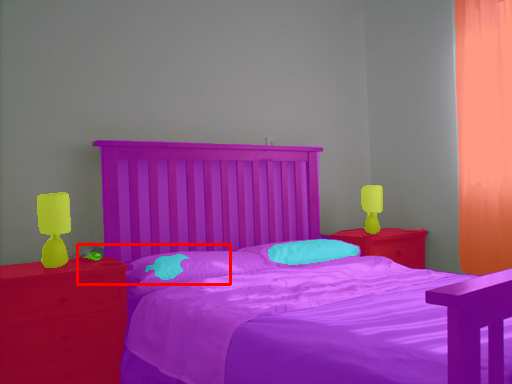}
\end{minipage}	
\begin{minipage}{0.2\textwidth}	
	\includegraphics[width=1\textwidth]{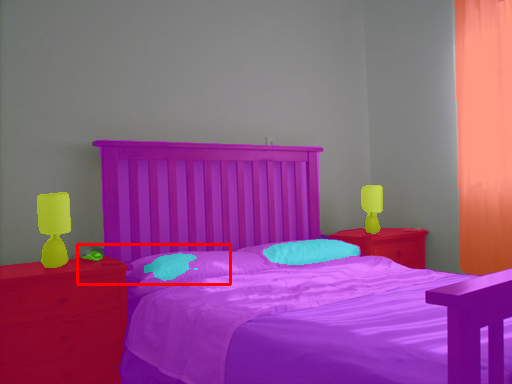}    	
\end{minipage}
\begin{minipage}{0.2\textwidth}
	\includegraphics[width=1\textwidth]{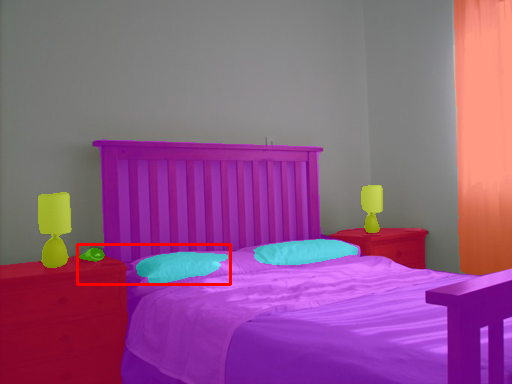}
\end{minipage}

\begin{minipage}{0.2\textwidth}
	\includegraphics[width=1\textwidth]{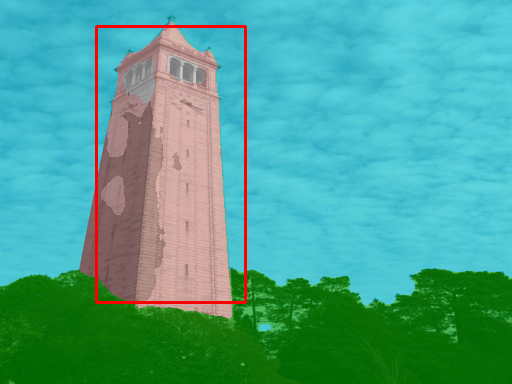}
\end{minipage}
\begin{minipage}{0.2\textwidth}
	\includegraphics[width=1\textwidth]{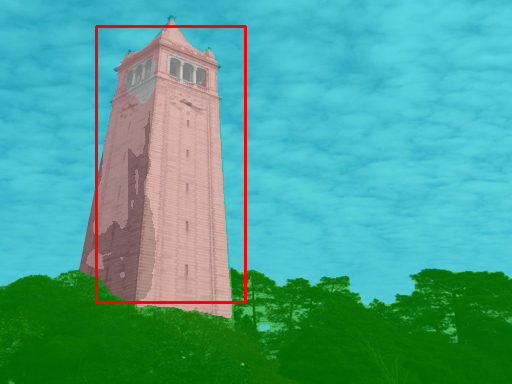}
\end{minipage}
\begin{minipage}{0.2\textwidth}	
	\includegraphics[width=1\textwidth]{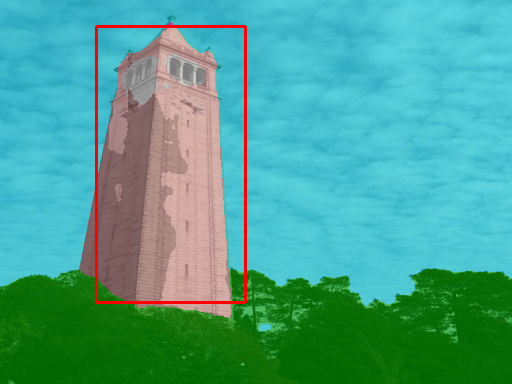}    	
\end{minipage}
\begin{minipage}{0.2\textwidth}
	\includegraphics[width=1\textwidth]{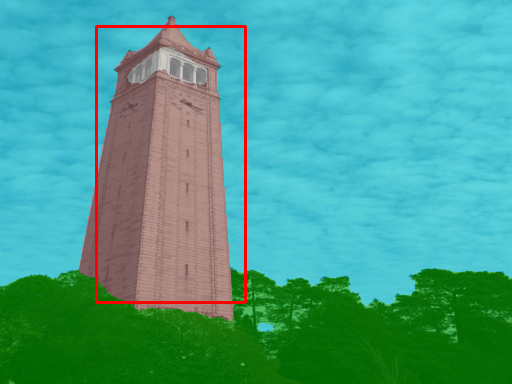}
\end{minipage}

\begin{minipage}{0.2\textwidth}
	\includegraphics[width=1\textwidth]{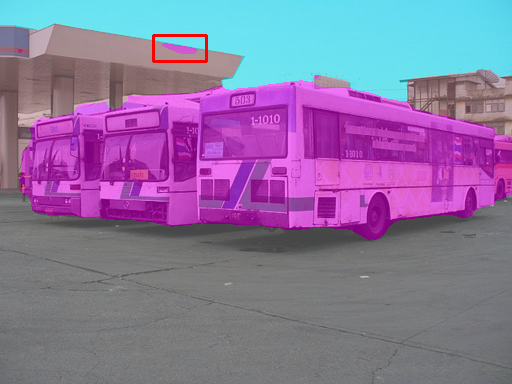}
\end{minipage}
\begin{minipage}{0.2\textwidth}
	\includegraphics[width=1\textwidth]{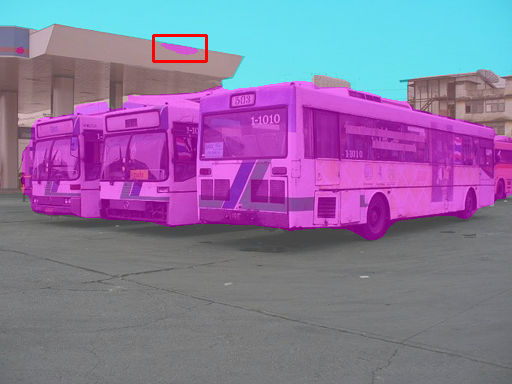}
\end{minipage}
\begin{minipage}{0.2\textwidth}	
	\includegraphics[width=1\textwidth]{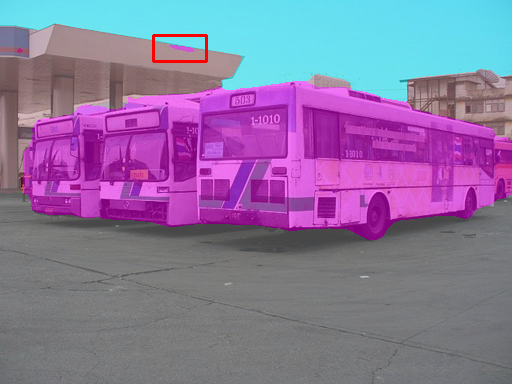}    	
\end{minipage}
\begin{minipage}{0.2\textwidth}
	\includegraphics[width=1\textwidth]{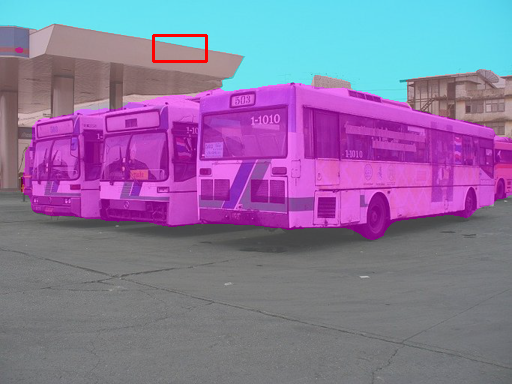}
\end{minipage}

\begin{minipage}{0.2\textwidth}\centering
	\includegraphics[width=1\textwidth]{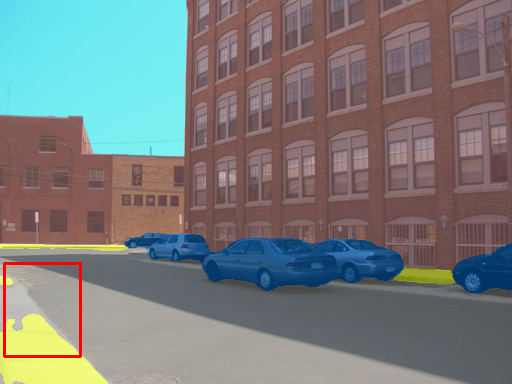}
(a) Cross-entropy 
\end{minipage}	
\begin{minipage}{0.2\textwidth}\centering	
	\includegraphics[width=1\textwidth]{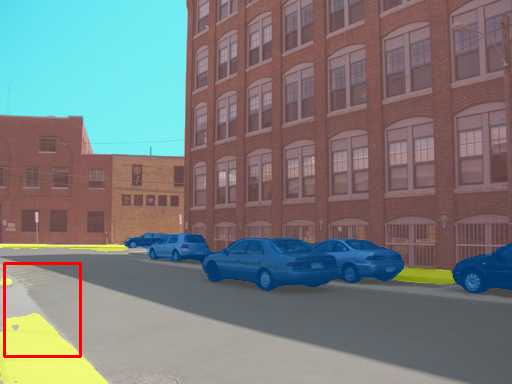}
(b) Focal loss 
\end{minipage}
\begin{minipage}{0.2\textwidth}\centering	
	\includegraphics[width=1\textwidth]{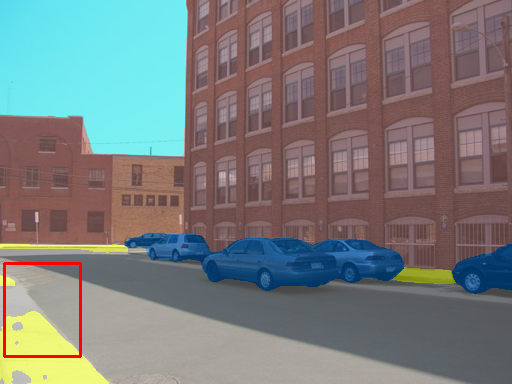} 
(c) Lov{\'a}sz-softmax
\end{minipage}
\begin{minipage}{0.2\textwidth}\centering	
	\includegraphics[width=1\textwidth]{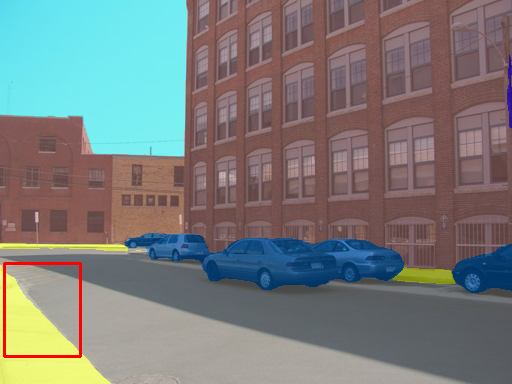}
(d) Margin calibration
\end{minipage}	
\caption{Scene parsing examples on MIT SceneParse150 validation set.}
\label{FIG:ADE20K}
\end{figure*}

The comparisons on the validation set with the three baseline methods are reported in Table \ref{TB:ADE20K}. Results show that when fixing the deep neural network, just using the proposed margin calibration method as a single learning objective, can boost the mIoU in very complex scene parsing tasks. When applying the flipping and multi-scale prediction, fine-tuning with focal loss and Lov{\'a}sz-softmax can improve the mIoU by 0.7\% and 0.5\%, respectively, while using the proposed margin calibration can achieve a further 0.9\% of performance gain compared to categorical cross-entropy. Some example segmentation results in both indoor and outdoor environments are illustrated in Fig. \ref{FIG:ADE20K}. We can see compared to other learning objectives, applying margin calibration can better annotate the \emph{pillow, building exterior} and {\em sidewalk} in these examples.

\begin{table}[t]
\centering
\caption{The segmentation performance on the MIT SceneParse150 validation set.}
\label{TB:ADE20K}
\begin{tabular}{|c|c|c|c|}
\hline
{\bf Method}	&Flipping & Multi-scale & mIoU   \\
\hline
 				& 			& & 37.7 \\
Cross-entropy 	&\cmark & & 37.9  \\
 				&\cmark	&\cmark & 39.1   \\
\hline
 				& 			& & 37.9  \\
Focal loss 		&\cmark & & 38.4  \\
 				&\cmark	&\cmark & 39.8  \\
\hline
 				& 			& & 37.9 \\
Lov{\'a}sz-softmax &\cmark & & 38.3  \\
 				&\cmark	&\cmark & 39.6  \\
\hline
 				& 			& & 38.5  \\
Margin calibration &\cmark & & 38.8  \\
 				&\cmark	&\cmark & {\bf 40.2}  \\
\hline

\end{tabular}
\end{table} 

\subsection{Results on Cityscapes, BDD100K, and Mapillary Vistas datasets}

\begin{table}[t]
\centering
\caption{The segmentation performance on the Cityscapes validation set (fine labels only).}
\label{TB:CITYSCAPES}
\begin{tabular}{|c|c|c|c|}
\hline
{\bf Method}	&Flipping & Multi-scale & mIoU   \\
\hline
 				& 			& & 78.7  \\
Cross-entropy 	&\cmark & & 78.9  \\
 				&\cmark	&\cmark & 79.4  \\
\hline
 				& 			& & 78.9 \\
Focal loss 		&\cmark & & 79.3  \\
 				&\cmark	&\cmark & 80.6  \\
\hline
 				& 			& & 79.6 \\
Lov{\'a}sz-softmax &\cmark & & 79.9  \\
 				&\cmark	&\cmark & 80.5  \\
\hline
 				& 			& & 80.0  \\
Margin calibration &\cmark & & 80.2  \\
 				&\cmark	&\cmark & {\bf 81.1}  \\
\hline

\end{tabular}
\end{table} 

\begin{table}[t]
\centering
\caption{The segmentation performance on the BDD100K validation set.}
\label{TB:BDD100K}
\begin{tabular}{|c|c|c|c|}
\hline
{\bf Method}	&Flipping & Multi-scale & mIoU   \\
\hline
 				& 			& & 64.4  \\
Cross-entropy 	&\cmark & & 64.6  \\
 				&\cmark	&\cmark & 65.7  \\
\hline
 				& 			& & 64.5  \\
Focal loss 		&\cmark & & 64.5  \\
 				&\cmark	&\cmark & 65.8  \\
\hline
 				& 			& & 64.6 \\
Lov{\'a}sz-softmax &\cmark & & 64.6  \\
 				&\cmark	&\cmark & 65.8  \\
\hline
 				& 			& & 64.8   \\
Margin calibration &\cmark & & 65.0  \\
 				&\cmark	&\cmark & {\bf 66.1}  \\
\hline
\end{tabular}
\end{table} 

\begin{table}[t]
\centering
\caption{The segmentation performance on the Mapillary Vistas validation set.}
\label{TB:MAPILLARY}
\begin{tabular}{|c|c|c|c|}
\hline
{\bf Method}	&Flipping & Multi-scale & mIoU   \\
\hline
 				& 			& & 49.1  \\
Cross-entropy 	&\cmark & & 49.3  \\
 				&\cmark	&\cmark & 49.8  \\
\hline
 				& 			& & 49.8  \\
Focal loss 		&\cmark & & 49.9  \\
 				&\cmark	&\cmark & 50.6  \\
\hline
 				& 			& & 49.7 \\
Lov{\'a}sz-softmax &\cmark & & 49.8  \\
 				&\cmark	&\cmark & 50.2  \\
\hline
 				& 			& & 50.2   \\
Margin calibration &\cmark & & 50.4  \\
 				&\cmark	&\cmark & {\bf 51.1}  \\
\hline
\end{tabular}
\end{table} 

We experimented the DeepLab v3+ model with different learning objectives on three street-view datasets. Similar to the training on the MIT SceneParse150 dataset, we fine-tuned the network parameters based on the pre-trained model obtained from the best checkpoint using categorical cross-entropy. The models on these datasets were independently trained. On BDD100K and Mapillary Vistas datasets, the images were re-scaled to $1280\times 720$ with the crop-size $720\times 720$. On the Cityscapes dataset, the images were not re-scaled but with the crop-size $800\times 800$. The ground-truth of test images of the three datasets are withheld by the organizers. However, the model performance of Cityscapes can be tested by submitting the segmentation results to their evaluation server.

Table \ref{TB:CITYSCAPES}, \ref{TB:BDD100K} and \ref{TB:MAPILLARY} show that using focal loss, Lov{\'a}sz-softmax and margin calibration can all lead to higher mIoU scores on the three validation sets. Specifically, using margin calibration to fine-tune the pretraind segmentation model can beat the second-best by 0.5\%, 0.3\%, 0.5\% on Cityscapes, BDD100K and Mapillary Vistas datasets, respectively. Focal loss is essentially a kind of dynamically scaled cross-entropy, where the scaling factor decays to zero as confidence in the correct class increases. This scaling factor can automatically down-weight the contribution of easy pixels during training and rapidly focus the model on hard pixels. So focal loss can generally replace cross-entropy in dense classification tasks. In our case, focal loss has a similar performance with Lov{\'a}sz-softmax, which is specifically designed for IoU optimization. However, thanks to the theoretical guarantee of the error bound and the explicit consideration of label imbalance, our method can achieve even higher mIoU compared to focal loss and Lov{\'a}sz-softmax.

\begin{figure*}[t]\centering
\begin{minipage}{0.23\textwidth}
	\includegraphics[width=1\textwidth]{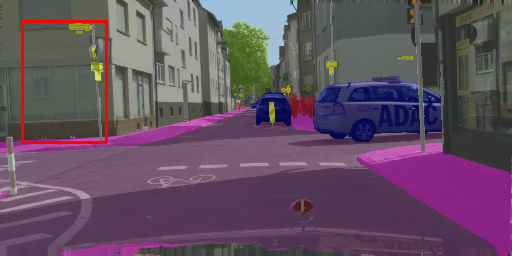}
\end{minipage}
\begin{minipage}{0.23\textwidth}
	\includegraphics[width=1\textwidth]{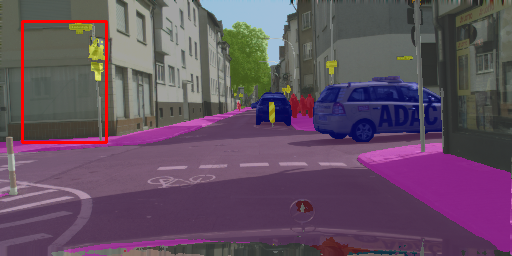}
\end{minipage}	
\begin{minipage}{0.23\textwidth}	
	\includegraphics[width=1\textwidth]{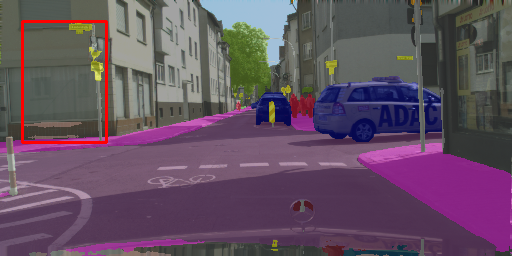}    	
\end{minipage}
\begin{minipage}{0.23\textwidth}
	\includegraphics[width=1\textwidth]{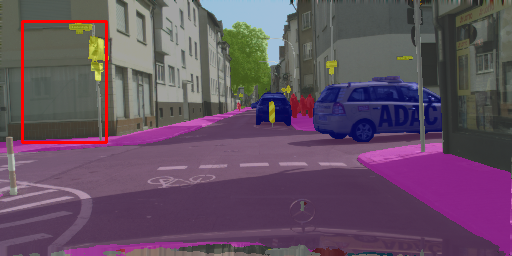}
\end{minipage}

\begin{minipage}{0.23\textwidth}
	\includegraphics[width=1\textwidth]{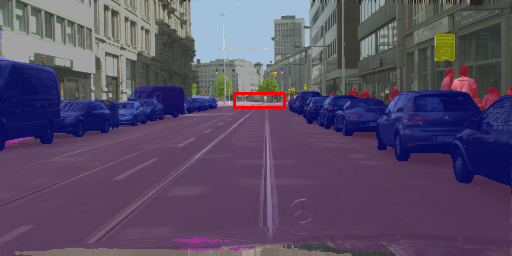}
\end{minipage}
\begin{minipage}{0.23\textwidth}
	\includegraphics[width=1\textwidth]{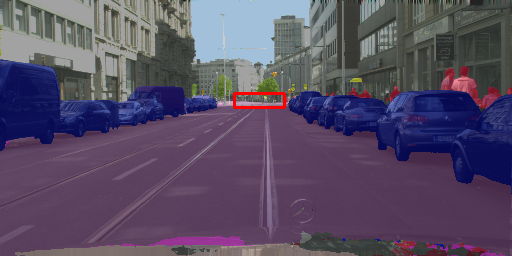}
\end{minipage}
\begin{minipage}{0.23\textwidth}	
	\includegraphics[width=1\textwidth]{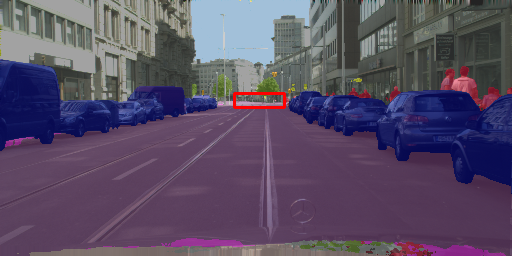}    	
\end{minipage}
\begin{minipage}{0.23\textwidth}
	\includegraphics[width=1\textwidth]{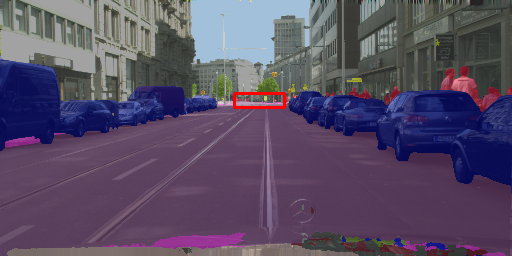}
\end{minipage}

\begin{minipage}{0.23\textwidth}
	\includegraphics[width=1\textwidth]{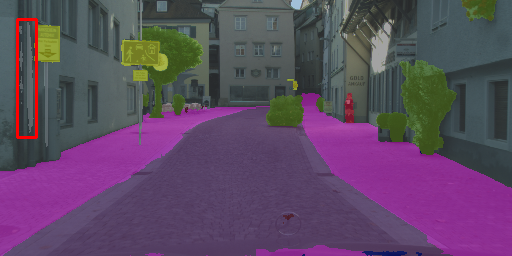}
\end{minipage}
\begin{minipage}{0.23\textwidth}
	\includegraphics[width=1\textwidth]{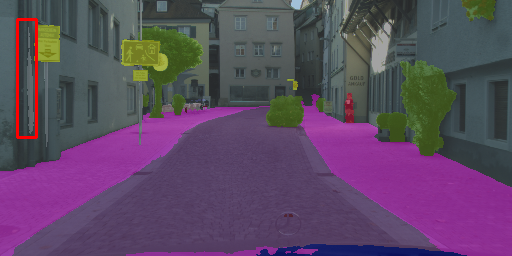}
\end{minipage}
\begin{minipage}{0.23\textwidth}	
	\includegraphics[width=1\textwidth]{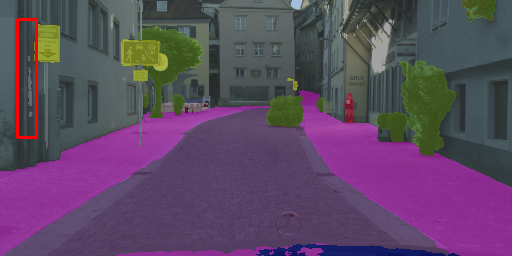}    	
\end{minipage}
\begin{minipage}{0.23\textwidth}
	\includegraphics[width=1\textwidth]{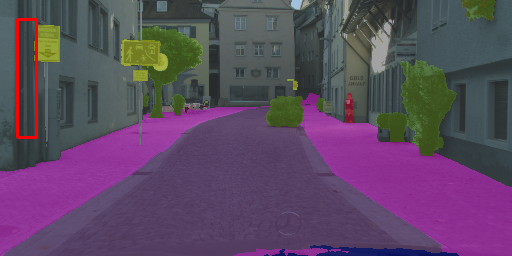}
\end{minipage}

\begin{minipage}{0.23\textwidth}\centering
	\includegraphics[width=1\textwidth]{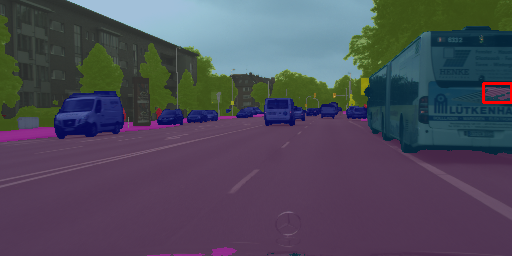}
(a) Cross-entropy 
\end{minipage}	
\begin{minipage}{0.23\textwidth}\centering	
	\includegraphics[width=1\textwidth]{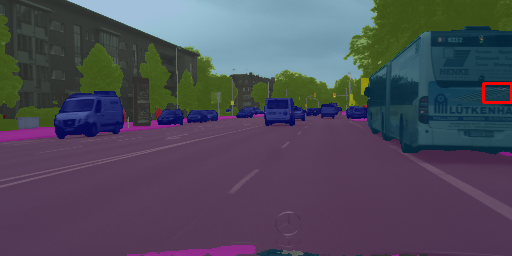}
(b) Focal loss 
\end{minipage}
\begin{minipage}{0.23\textwidth}\centering	
	\includegraphics[width=1\textwidth]{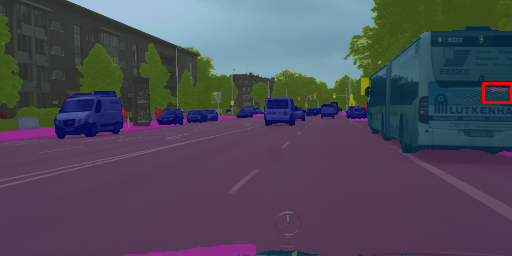} 
(c) Lov{\'a}sz-softmax
\end{minipage}
\begin{minipage}{0.23\textwidth}\centering	
	\includegraphics[width=1\textwidth]{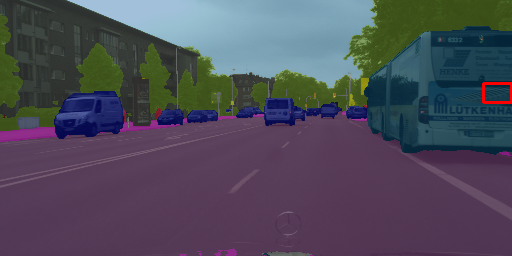}
(d) Margin calibration
\end{minipage}	
\caption{Semantic segmentation examples on the Cityscapes validation set.}
\label{FIG:CITYSCAPES}
\end{figure*} 

We submitted the prediction with defferent settings to the evaluation server of Cityscapes\footnote{\url{https://www.cityscapes-dataset.com/method-details/?submissionID=10089}}. The overall comparisons of our model with some recently proposed methods are summarized in Table \ref{TB:OVERALL-CITYSCAPES}. Note that our method mainly aims to optimize the mIoU measure named {\bf Class IoU} in Cityscapes. From the table, we can see that training with margin calibration, a single deep segmentation model achieves very promising results. Without the pre-training using the 20,000 coarsely labelled images, simply replacing the cross-entropy with the proposed margin calibration, the mIoU can be improved by 1\%. If the model is pre-traind with the coarsely labelled data then finetuned, the final mIoU can be further boosted by 0.6\%. Compared to the original implementation of DeepLab v3+ in \cite{ECCV18:DEEPLAB}, our segmentation model backend on SEResNeXt-50, which is a shallower network pre-traind on ImageNet-1K \cite{IJCV:IMAGENET} but not the much larger JFT-300M \cite{CVPR17:JFT}. Even so, with the margin calibration as a better learning objective, the final performance of our implementation is slightly better than the Deeplab v3+ backend on Aligned Xception. Some examplar segmentation results for the scene parsing visualization are illustrated in Fig. \ref{FIG:CITYSCAPES}. We can see that fine-tuning with margin calibration can generally reduce false positives and lead to finer details. 

\begin{table*}[t]
\centering 
\caption{Online evaluation of Cityscapes testing set.}
\label{TB:OVERALL-CITYSCAPES}
\begin{tabular}{|c|c|c|cccc|}
\hline
{\bf Method} &Backbone & Coarse data	&IoU cla. &iIoU cla. &IoU cat. &iIoU cat.  \\
\hline
AAF \cite{ECCV18:AAF}	& ResNet-101 & \xmark &79.1 &56.1 &90.8 &78.5 \\
PSANet\cite{ECCV18:PSANET} & ResNet-101 & \cmark  &80.1 &- &- &- \\
Pad-Net\cite{CVPR18:PAD_NET} & ResNet-101 & \xmark &80.3 &58.8 &90.8 &78.5 \\
SPGNet\cite{ICCV19:SPGNET} & $2\times$ResNet-50 & \xmark &81.1 &61.4 &92.1 &82.1 \\
PSPNet\cite{CVPR17:PSPNET} & ResNet-101 & \cmark	&81.2 &59.6 &91.2 &79.2   \\
L2-SP\cite{ICML18:L2SP} & ResNet-101 & \cmark &81.2 &58.1 &91.0 &78.5 \\
DANet\cite{CVPR19:DANET} & ResNet-101 & \xmark &81.5 &62.3 &91.6 &82.6 \\
DeepLab v3+ \cite{ECCV18:DEEPLAB} & Aligned Xception & \cmark &{\bf 82.1} &62.4 &92.0 &81.9   \\
HANet \cite{CVPR20:HANET} & ResNet-101 &  \xmark &80.9 &58.6 &91.2 &79.5 \\
HRNetV2 \cite{TPAMI:HRNET} & HRNetV2-W48 & \xmark &81.8 &61.2 &{\bf 92.2} &{\bf 82.1} \\
\hline
DeepLab v3+ (Cross-entropy)	& SEResNeXt-50 & \xmark &80.5 &58.3 &91.7 &80.8   \\
DeepLab v3+ (Focal loss)	& SEResNeXt-50 & \xmark &81.1 &59.7 &92.0 &81.5   \\
DeepLab v3+ (Lov{\'a}sz-softmax)	& SEResNeXt-50 & \xmark &81.0 &60.0 &91.8 &81.3   \\

DeepLab v3+ (Margin calibration)	& SEResNeXt-50 & \xmark &81.5 &59.9 &91.9 &81.1   \\

DeepLab v3+ (Margin calibration)	& SEResNeXt-50 & \cmark &{\bf 82.1}  &{\bf 62.5} &92.1 &81.8   \\

\hline
\end{tabular}
\end{table*}

\section{Conclusion} \label{SEC:CONCLUSION}
We have presented a versatile distribution-aware margin calibration method as a better learning method, to optimize the Jaccard index in image semantic segmentation. With the consideration of both empirical performance and the error bound, the scheme can increase the discriminative power with a better generalization ability. We gave both theoretical and experimental analysis to demonstrate its effectiveness, substantially improving the IoU scores by inserting it into a deep semantic segmentation network. 

\bibliographystyle{named}
\bibliography{reference} 

\appendix

\section{Appendix}

\subsection{Proof of Theorem 1}

We first prove with probability $1-\frac{\eta}{K}$, the following inequality holds:
\begin{equation}
\mathpzc{IoU}_k \geq \overline{IoU}_k - \epsilon_k.
\end{equation} 

Assume the following inequality holds for non-negative $\epsilon_{k0}$ and $\epsilon_{0k}$:
\begin{equation} \label{ub}
\mathpzc{IoU}_k  =  \frac{\mathpzc{P}_k - \mathpzc{P}_{k0} }{ \mathpzc{P}_k + \mathpzc{P}_{0k} } \geq 
\frac{\mathpzc{P}_k - (\mathpzc{P}_{k0} -\epsilon_{k0})}{ \mathpzc{P}_k + (\mathpzc{P}_{0k} -\epsilon_{0k})} - \epsilon_k.
\end{equation}
Solving the above inequality, we can get:
\begin{equation} \label{finalepslk}
\epsilon_k = (\frac{a_k}{b_k}\epsilon_{0k}+\epsilon_{k0})(b_k-\epsilon_{0k})^{-1},
\end{equation}
where $a_k=\mathpzc{P}_k-\mathpzc{P}_{k0}$ and $b_k=\mathpzc{P}_k+\mathpzc{P}_{0k}$.

Next, we should get the values of $\epsilon_{k0}$ and $\epsilon_{0k}$ to calculate $\epsilon_k$, which should satisfy the following inequality:
\begin{equation}\label{ub2}
\frac{\mathpzc{P}_k - (\mathpzc{P}_{k0}-\epsilon_{k0})}{\mathpzc{P}_k + (\mathpzc{P}_{0k}-\epsilon_{0k})} \geq \overline{IoU}_k = \frac{P_k - \ell_{k0}(\theta,\rho_{k0})}{P_k + \ell_{0k}(\theta,\rho_{0k})},
\end{equation}
so we can simply substitute (\ref{ub2}) into (\ref{ub}) to complete the proof.

The empirical label distribution $P_k$ is irrelevant to the model $\theta$ and we assume it is an accurate estimation of the  label distribution $\mathcal{D}_{\mathcal{Y}}$, i.e.,
\begin{equation}
\mathpzc{P}_k=\mathbf{P}_{y\sim \mathcal{D}_{\mathcal{Y}}}(y=k)\approx P_k  = \frac{1}{N}\sum\limits_{i=1}^{N}\mathbb{I}(y_i=k).
\end{equation}

Based on the above approximation, a sufficient condition for (\ref{ub2}) regarding $\epsilon_{0k}$ and $\epsilon_{k0}$ is:
\begin{align}\label{eps2}
\mathpzc{P}_{k0} -\epsilon_{k0} &\leq \ell_{k0}(\theta,\rho_{k0}),  \nonumber\\
\mathpzc{P}_{0k} -\epsilon_{0k} &\leq \ell_{0k}(\theta,\rho_{0k}) 
\end{align}
Following the margin-based generalization bound in \cite[Theorem 9.2]{mohri2018foundations}, for the $N_k$ foreground class pixels, with the probability at least $1-\frac{\eta}{2K}$, we have:
\begin{equation}\label{gb}
\mathpzc{P}_{k0} - \ell_{k0}(\theta,\rho_{k0}) \leq \frac{N_k}{N}(\frac{4K}{\rho_{k0}}\mathfrak{R}_{N_k}(\Theta)+\sqrt{\frac{2M\log \frac{2K}{\eta}}{N_k}}),
\end{equation}
where $\mathfrak{R}_{N_k}(\Theta)$ is the Rademacher complexity for the hypothesis class $\Theta$ over the $N_k$ pixels of the $k$-th foreground class. For an input data batch with $M$ pixels, we first apply the McDiarmid's inequality for $M$-dependent data \cite{liu2019mcdiarmid} to the proof of \cite[Theorem 3.3]{mohri2018foundations}. Then we use it in the proof of \cite[Theorem 9.2]{mohri2018foundations} to get the formulation of (\ref{gb}). 

The Rademacher complexity $\mathfrak{R}_{N_k}(\Theta)$ typically scales in $\sqrt{\frac{C(\Theta)}{N_k}}$ with $C(\Theta)$ being the a proper complexity measure of $\Theta$ \cite{neyshabur2018role}, and such a scale has also been used in related work (see \cite{cao2019learning} and the references therein). We can then rewrite (\ref{gb}) as:
\begin{equation}\label{epk0}
\mathpzc{P}_{k0}(\theta) - \ell_{k0}(\theta,\rho_{k0}) \leq \frac{\sqrt{N_k}}{N} \frac{4K}{\rho_{k0}}\mathcal{F},
\end{equation}
where $\sigma(\frac{1}{\eta}) \triangleq \frac{\rho_{\max}}{4K} \sqrt{2M\log \frac{2K}{\eta}}$ is typically a low-order term in $\frac{1}{\eta}$. 

Similarly, for the $N-N_k$ pixels of the background class, with the probability at least $1-\frac{\eta}{2}$,
\begin{equation}\label{ep0k}
\mathpzc{P}_{0k}(\theta) - \ell_{0k}(\theta,\rho_{0k}) \leq \frac{\sqrt{N-N_k}}{N}\frac{4K}{\rho_{0k}} \mathcal{F}
\end{equation}
for the $N-N_k$ background class pixels.

We then combine (\ref{epk0}), (\ref{ep0k}), (\ref{eps2}) and take a union bound over $\epsilon_{k0}$ and $\epsilon_{0k}$, to get following equations, with which (\ref{eps2}) holds with the probability at least $1-\eta/K$:
\begin{align}
\epsilon_{k0}&= \frac{\sqrt{N_k}}{N} \frac{4K}{\rho_{k0}}\mathcal{F},\nonumber \\
\epsilon_{0k}&= \frac{\sqrt{N-N_k}}{N}\frac{4K}{\rho_{0k}} \mathcal{F}.
\end{align}

Then we substitute above equations into (\ref{finalepslk}). Let $\mu_k = \frac{\rho_{k0}}{\rho_{0k}}$, we have:
\begin{equation}\label{bfapp}
\epsilon_k = \frac{\frac{a_k}{b_k}\sqrt{N-N_k} + \frac{\sqrt{N_k}}{\mu_k}} {\frac{b_k N}{4K\mathcal{F}}\rho_{0k}-\sqrt{N-N_k}},
\end{equation} 
so that with the probability at least $1-\eta/K$ the inequality (\ref{tobepf}) holds. 

In practice, we do not know the values of $a_k$ and $b_k$ so that Eq.(\ref{bfapp}) has its own limitations. However, we know $\frac{a_k}{b_k}\leq 1$ and $b_k\geq \mathpzc{P}_k$ so we can get a very useful bound:
\begin{equation}\label{aftapp}
\epsilon_k \leq \frac{\sqrt{N-N_k} + \frac{\sqrt{N_k}}{\mu_k}} {\frac{N_k}{4K\mathcal{F}}\rho_{0k}-\sqrt{N-N_k}}.
\end{equation}

Averaging the union bound Eq.(\ref{tobepf}) over all classes, we can obtain the following inequality with probability at least $1-\eta$:
\begin{equation}
\mathpzc{mIoU}\ge \overline{mIoU}-\epsilon,
\end{equation}
with
\begin{equation}\label{finalresul}
\epsilon = \frac{1}{K}\sum\limits_{k=1}^{K} \frac{ \sqrt{N-N_k}+  \frac{\sqrt{N_k}}{\mu_k}}{\frac{N_k}{4K\mathcal{F}}\rho_{0k}-\sqrt{N-N_k}},
\end{equation}
where we complete the proof.

\subsection{Proof of Colollary 1}

We substitute $\mu_k$ in (\ref{finalresul}) with $\frac{\sqrt{N_k}}{r(N/N_k\!-\!1)-\sqrt{N\!-\!N_k}}$, where $r$ is a hyper-parameter, we can get:
\begin{align} 
\epsilon 
& = \frac{1}{K}\sum_{k=1}^{K} (\frac{r(N-N_k)}{N_k})(\frac{N_k}{4K\mathcal{F}}\rho_{0k}-\sqrt{N-N_k})^{-1} \nonumber\\
& = \frac{1}{K}\sum_{k=1}^{K} (\frac{r(N-N_k)}{N_k^2})(\frac{1}{4K\mathcal{F}}\rho_{0k}-\frac{\sqrt{N-N_k}}{N_k})^{-1}.
\end{align}
Let $x_k = \frac{r(N-N_k)}{N_k^2} $ and $y_k = \frac{1}{4K\mathcal{F}}\rho_{0k}-\frac{\sqrt{N-N_k}}{N_k}$. According to Cauchy-Schwarz inequality we have:
\begin{equation}
\left(\sum_{k=1}^{K} \sqrt\frac{x_k}{y_k}\cdot \sqrt{y_k}\right)^2 \leq (\sum_{k=1}^{K}\frac{x_k}{y_k}) (\sum_{k=1}^{K} y_k),
\end{equation}
so that
\begin{align}
\epsilon &\geq \frac{1}{K}\cdot\left(\sum\limits_{k=1}^{K} \sqrt{x_k}\right)^2 (\sum\limits_{k=1}^{K} y_k)^{-1} \nonumber \\
&= \frac{r}{K}\cdot\frac{\left(\sum\limits_{k=1}^{K} \frac{\sqrt{N-N_k}}{N_k}\right)^2}{\frac{1}{4K\mathcal{F}}\sum\limits_{k=1}^{K} \rho_{0k}- \sum\limits_{k=1}^{K}\frac{\sqrt{N-N_k}}{N_k}}. 
\end{align}

The RHS of the equality is a constant because $r$ is a given hyper parameter and we assume  $\sum\limits_{k=1}^{K} \rho_{0k} = \text{some contant}$. The equality holds when $\frac{\sqrt{x_1}}{y_1}=\ldots=\frac{\sqrt{x_K}}{y_K}$, which yields Corollary \ref{proprho}. 

Note that $\mu_k = \frac{\sqrt{N_k}}{r(N/N_k-1)-\sqrt{N-N_k}}$, while in Corollary \ref{proprho} $\mu_k = \frac{P_k\sqrt{N_k}}{\upsilon(N-N_k)-P_k\sqrt{N-N_k}}$. These two conditions are essentially equivalent when $r$ and $\upsilon$ are hyper-parameters. To see this, simply let $r=N\upsilon$ and notice that $P_k=\frac{N_k}{N}$.

\end{document}